\documentclass{article}
\usepackage{ijcai24}

\usepackage{times}
\usepackage{soul}
\usepackage{url}
\usepackage[hidelinks]{hyperref}
\usepackage[utf8]{inputenc}
\usepackage[small]{caption}
\usepackage{graphicx}
\usepackage{amsmath}
\usepackage{amsthm}
\usepackage{booktabs}
\usepackage{algorithm}
\usepackage{algorithmic}
\usepackage[switch]{lineno}

\usepackage{amssymb,amsfonts}
\usepackage{subfigure}
\usepackage{tabularx}
\usepackage{makecell}

\title{Benchmarking Stroke Forecasting with Stroke-Level Badminton Dataset}

\author{
    Wei-Yao Wang, Wei-Wei Du, Wen-Chih Peng, Tsi-Ui Ik
    \affiliations
    Department of Computer Science, National Yang Ming Chiao Tung University
    \emails
    sf1638.cs05@nctu.edu.tw, \{wwdu.cs10, wcpengcs, tik\}@nycu.edu.tw
}

\begin{document}

\maketitle

\begin{abstract}
In recent years, badminton analytics has drawn attention due to the advancement of artificial intelligence and the efficiency of data collection.
While there is a line of effective applications to improve and investigate player performance, there are only a few public badminton datasets that can be used by researchers outside the badminton domain.
Existing badminton singles datasets focus on specific matchups; however, they cannot provide comprehensive studies on different players and various matchups.
In this paper, we provide a badminton singles dataset, ShuttleSet22, which is collected from high-ranking matches in 2022.
ShuttleSet22 consists of 30,172 strokes in 2,888 rallies in the training set, 1,400 strokes in 450 rallies in the validation set, and 2,040 strokes in 654 rallies in the testing set, with detailed stroke-level metadata within a rally.
To benchmark existing work with ShuttleSet22, we hold a challenge, Track 2: Forecasting Future Turn-Based Strokes in Badminton Rallies, at CoachAI Badminton Challenge @ IJCAI 2023, to encourage researchers to tackle this real-world problem through innovative approaches and to summarize insights between the state-of-the-art baseline and improved techniques, exchanging inspiring ideas.
The baseline codes and the dataset are made available at the GitHub repo\footnote{https://github.com/wywyWang/CoachAI-Projects/tree/main/CoachAI-Challenge-IJCAI2023}.
% we test the state-of-the-art stroke forecasting approach, ShuttleNet, with the corresponding stroke forecasting task, i.e., predict the future strokes based on the given strokes of each rally.
% Moreover, we hold a challenge, Track 2: Forecasting Future Turn-Based Strokes in Badminton Rallies, at CoachAI Badminton Challenge @ IJCAI 2023 to encourage researchers to tackle this real-world problem through different approaches.

\end{abstract}
\section{Introduction}

Sports analytics has garnered increasing attention since the advancement of technology, which greatly facilitates data collection and increases data diversity.
In recent years, there has been a surge in studies applying advanced artificial intelligence techniques, e.g., computer vision on video frames \cite{DBLP:conf/kdd/KimKCYK22}, and machine learning models for action valuing \cite{DBLP:conf/kdd/DecroosBHD19,DBLP:conf/kdd/MerhejBMR21}.
Badminton, as one of the major racket sports worldwide in terms of participation, demands high physical and tactical conditions, attracting researchers to introduce novel applications \cite{Wang2020badminton}.
For instance, \cite{10.1145/3551391} propose long short-term extractors to quantify the win influence of each shot within a rally, while \cite{DBLP:journals/corr/abs-2211-12217} design the movement forecasting task to predict players' movements using graph-based approaches.

In this paper, our aim is to introduce, ShuttleSet22, a stroke-level badminton singles dataset collected from real-world high-ranking matches in 2022.
ShuttleSet22 extends the original ShuttleSet \cite{ShuttleSet}, comprising 30,172 strokes (2,888 rallies), 1,400 strokes (450 rallies) in the validation set, and 2,040 strokes (654 rallies) in the testing set.
While ShuttleSet22 shares similar stroke-level data formats with ShuttleSet, it consists of matches in 2022 instead of the period between 2018 and 2021.
Therefore, ShuttleSet22 can be considered the most recent iteration of the badminton singles dataset, enabling the examination of model effectiveness in recent matches.
ShuttleSet22 is sourced from public videos\footnote{http://bwf.tv/} and has been meticulously labeled by domain experts with the shot-by-shot labeling tool \cite{DBLP:conf/apnoms/HuangHLIW22}.

\begin{table}
    \centering
    \small
    \caption{Comparison of the previous badminton dataset with our proposed ShuttleSet22.}
    \label{tab:datasets_comparison}
    \scalebox{0.80}{\begin{tabular}{c|ccccc}
    \toprule
     & \# Players & \# Match & \# Rally & \# Stroke & Year\\
    \midrule
    \makecell[l]{BadmintonDB\\ \cite{DBLP:conf/mm/BanSAL22}} & 2 & 9 & 811 & 9,671 & 2018-2020\\
    \midrule
    ShuttleSet22 (Ours) & 35 & 58 & 3,992 & 33,612 & 2022\\
    \bottomrule
\end{tabular}}
\end{table}

To boost researchers' engagements in badminton analytics, we have, for the first time, initiated a challenge within CoachAI Badminton Challenge 2023\footnote{https://sites.google.com/view/coachai-challenge-2023/} in conjunction with IJCAI 2023.
Specifically, we have organized the forecasting of future turn-based strokes in badminton rallies (Track 2) and provided the state-of-the-art stroke forecasting approach \cite{DBLP:conf/aaai/WangSCP22} as the official baseline.
This track has attracted approximately 100 participants aiming to improve the effectiveness of the stroke forecasting task, with 16 teams submitting their final results on the leaderboard.
In addition to the results, we summarize these insights and analyses of this challenge to inspire ideas for bridging the gap between badminton analytics and artificial intelligence communities\footnote{The video can be found at https://youtu.be/yhRouMpxb2M.}.

% The paper is organized as follows: Related work is described in Section 2, the details of the stroke forecasting task, the data exploration, and the official baseline are introduced in Section 3, methods of participants are summarized in Section 4, the results from participants are depicted in Section 5, and we conclude our findings in Section 6.
% We select it as the official baseline due to not only the state-of-the-art performance but also the first work for turn-based sequence prediction.
\begin{table*}[t]
    \small
    \centering
    \caption{The statistics of shot types. Short and long services only happen at the first stroke.}
    \label{tab:shot_type}
    \scalebox{0.96}{
        \begin{tabular}{c||c|c||c|c||c|c||c}
    \toprule
    Shot Type & Not Around Head & Around Head & Forehand & Backhand & Higher than the net & Below than the net & Total Count\\
    \midrule
    Drive & 891 & 41 & 701 & 231 & 497 & 435 & 932 \\
    Net Shot & 5721 & 4 & 3110 & 2615 & 158 & 5566 & 5725 \\
    Lob & 5205 & 2 & 2706 & 2501 & 4890 & 316 & 5207 \\
    Clear & 2062 & 1016 & 2808 & 270 & 2868 & 210 & 3078 \\
    Drop & 2539 & 643 & 2863 & 319 & 133 & 3049 & 3182 \\
    Push/Rush & 1994 & 17 & 992 & 1019 & 1277 & 734 & 2011 \\
    Smash & 2402 & 1280 & 3641 & 41 & 56 & 3625 & 3682 \\
    Defensive Shot & 4079 & 8 & 1742 & 2345 & 940 & 3146 & 4087 \\
    \midrule
    Short Service & 1620 & 0 & 146 & 1474 & 21 & 1599 & 1620 \\
    Long Service & 648 & 0 & 528 & 120 & 644 & 3 & 648 \\
    \bottomrule
\end{tabular}
    }
\end{table*}

\section{Related Work: Public Badminton Datasets}

Generally, there are only a few public badminton datasets due to the heavy cost of collecting and labeling fine-grained records by domain experts \cite{DBLP:conf/cikm/Wang22}.
Recently, researchers have released badminton datasets to foster the sports community.
For instance, the shuttlecock datasets \cite{shuttlecock-cqzy3_dataset} contains 8K images of shuttlecocks which are resized to 640*640 pixels. This dataset includes the position of shuttlecocks which could train an object detection model.
The BadmintonDB \cite{DBLP:conf/mm/BanSAL22} features rally, strokes, and outcome annotations between two players, which can be used in player-specific match analysis and prediction tasks, which is also the stroke-level dataset.
However, BadmintonDB only consists of the same matchup (i.e., Kento Momota and Anthony Sinisuka Ginting) instead of the various matchups.
Moreover, BadmintonDB collects matches from 2018 to 2020, while our ShuttleSet22 collects high-ranking matches in 2022 to reflect the state-of-the-art tactic records.
We summarize the discrepancies between our ShuttleSet and BadmintonDB in Table \ref{tab:datasets_comparison}.

\section{The Challenge System}

\begin{figure}
  \centering
  \includegraphics[width=\linewidth]{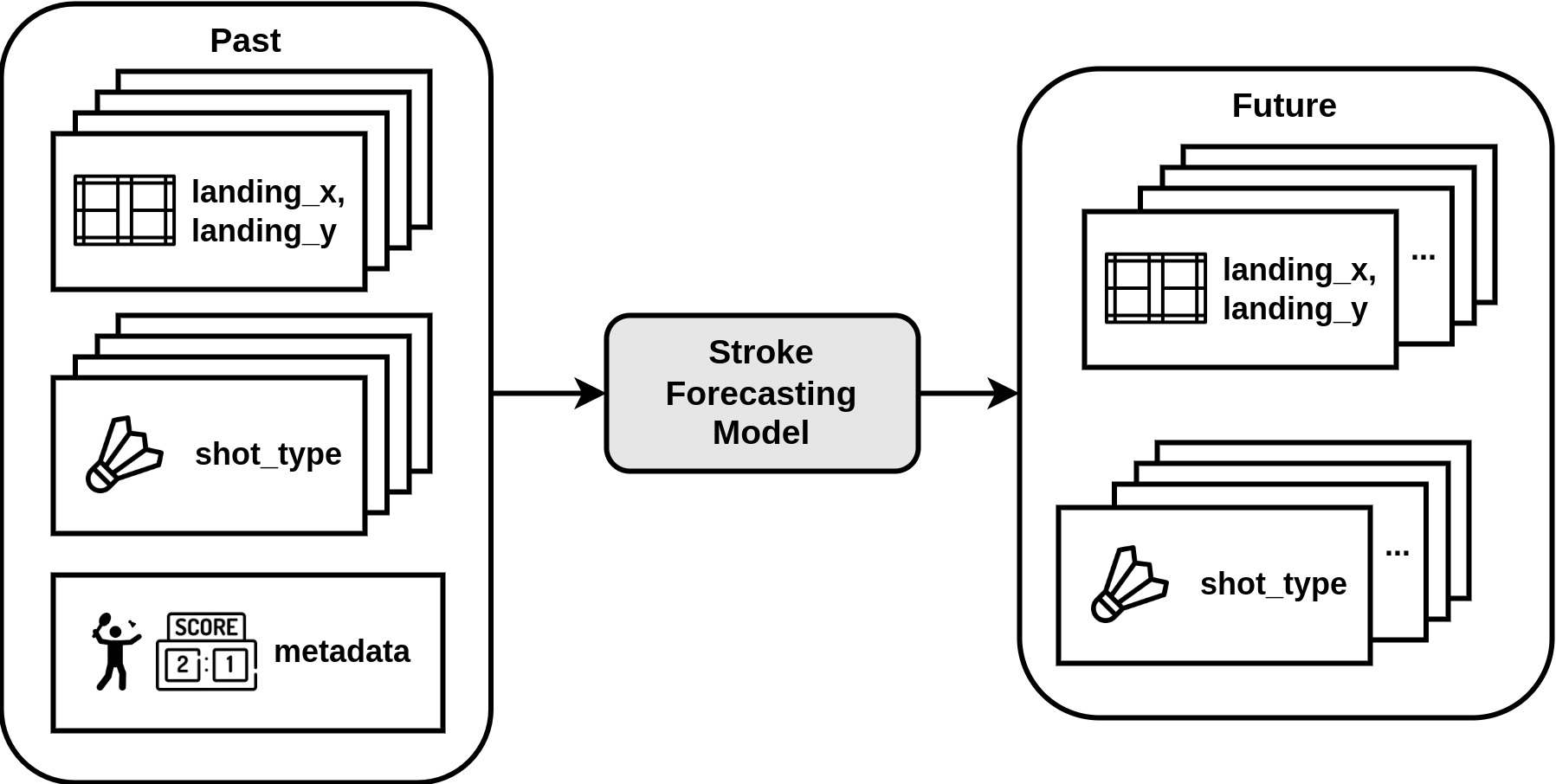}
  \caption{Illustrated system of Track 2: Forecasting Future Turn-Based Strokes in Badminton Rallies.}
  \label{fig:flowchart}
\end{figure}

Conventional applications in badminton mainly focus on quantifying stroke performance \cite{DBLP:conf/icdm/WangCYWFP21} or retrieving information from videos \cite{DBLP:conf/mir/ChuS17}, which motivates us to further investigate a more challenging yet critical real-world application.
The goal of this shared task is to \textbf{forecasting future turn-based strokes in badminton rallies}, which aims to design forecasting models capable of predicting future strokes, including shot types and locations, based on past stroke sequences.
Figure \ref{fig:flowchart} illustrates the entire flowchart of our task.
As we have built the pipeline from data processing to evaluations, participants only need to modify the stroke forecasting model to swiftly iterate their approaches.
The task page is available on \href{https://codalab.lisn.upsaclay.fr/competitions/12017?secret\_key=ba1acc46-1279-4781-94f4-0510fdb017ca}{\underline{Codalab}}.

\begin{figure}
  \centering
    \includegraphics[width=0.97\linewidth]{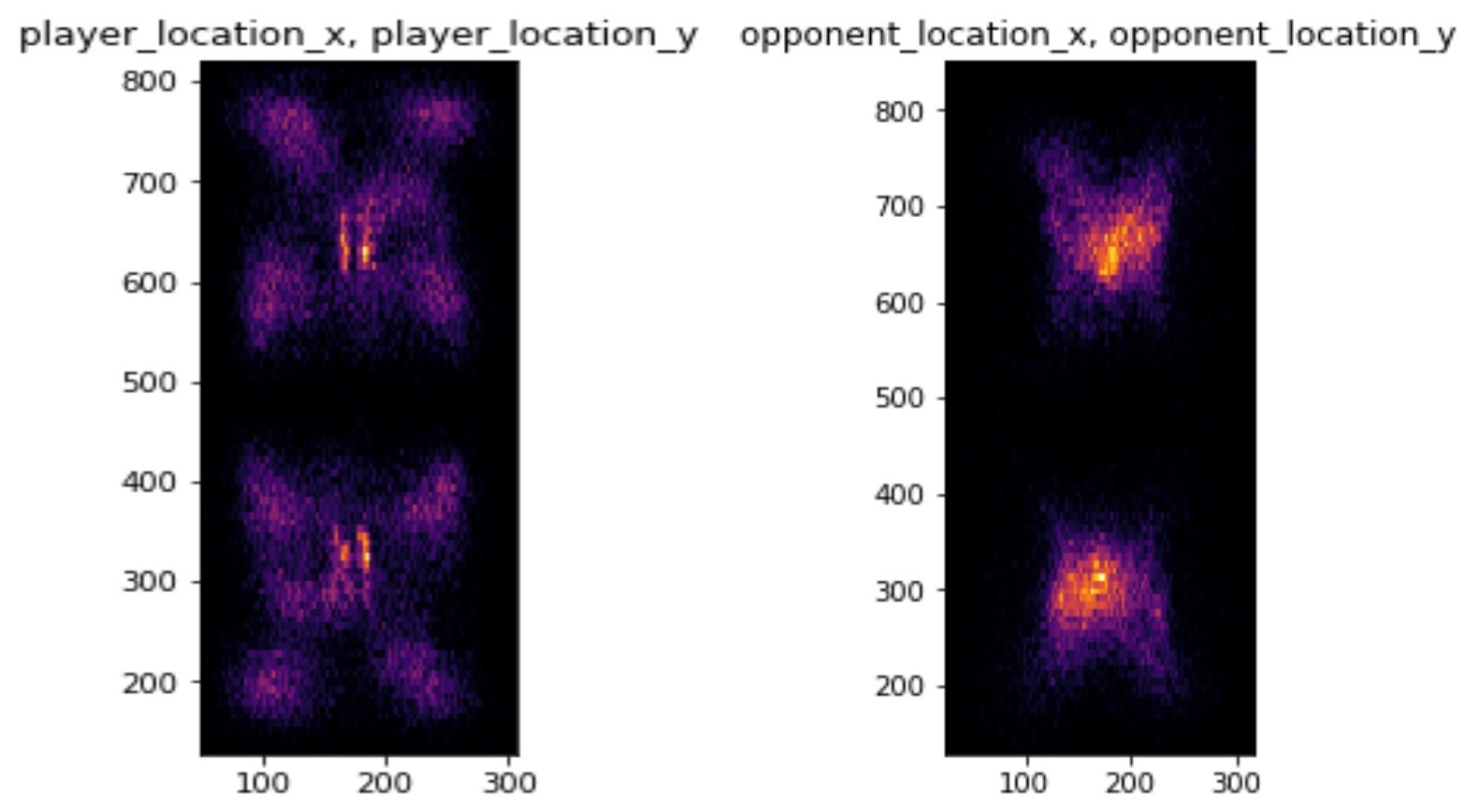}
  \caption{
  Heatmap of player and opponent locations in ShuttleSet22.
  }
  \label{fig:player_opponent_location}
\end{figure}

\subsection{Problem Formulation}
Following the definition outlined in \cite{DBLP:conf/aaai/WangSCP22}, for each singles rally, given the observed $\tau$ strokes with type-area pairs and two players, the goal is to predict the future strokes, including shot types and area coordinates, for the next $n$ steps. 
To simplify the problem, $\tau$ is set to be 4 in our challenge, and $n$ is various based on the length of the rally, which is given to the participants.

% take what you need from the information below and delete non-used (do not use all of them directly since it will have some copy issues.
Let $R=\{S_r, P_r\}_{r=1}^{|R|}$ denote historical rallies of badminton matches, where the $r$-th rally is composed of a stroke sequence with type-area pairs $S_r=(\langle s_1, a_1\rangle,\cdots,\langle s_{|S_r|}, a_{|S_r|}\rangle)$ and a player sequence $P_r=(p_1,\cdots,p_{|S_r|})$.
At the $i$-th stroke, $s_i$ represents the shot type, $a_i=\langle x_i, y_i\rangle \in \mathbb{R}^{2}$ are the coordinates of the shuttle destinations, and $p_i$ is the player who hits the shuttle. We denote Player A as the served player and Player B as the other for each rally in this paper. For instance, given a singles rally between Player A and Player B, $P_r$ may become $(A, B, \cdots, A, B)$.
We formulate the problem of stroke forecasting as follows. For each rally, given the observed $\tau$ strokes $(\langle s_i, a_i\rangle)_{i=1}^{\tau}$ with players $(p_i)_{i=1}^{\tau}$, the goal is to predict the future strokes including shot types and area coordinates for the next $n$ steps, i.e., $(\langle s_i, a_i\rangle)_{i={\tau+1}}^{\tau+n}$.
We note that $n$ is pre-defined as the actual length of the corresponding rally.

\subsection{Exploratory Data Analysis}
We have released 58 matches, approximately 4,000 rallies with shot-level records following the BLSR format \cite{10.1145/3551391} for this task, where parts of the matches from 2018 to 2021 are identical to those in \cite{DBLP:conf/aaai/WangSCP22}, while we have further included new matches collected from 2022.

%% Table of column explanation

% \begin{table*}
%     \small
%     \centering
%     \caption{Explanation and example of each column.}
%     \label{tab:column}
%     \input{table/data_columnname}
% \end{table*}

%% How we split the datasets

%% Some statistics of the datasets
% Table \ref{tab:column} summarizes the definition of each column, which is mainly followed by \cite{10.1145/3551391}.
Figure \ref{fig:player_opponent_location} shows the heatmap of player location and opponent location, indicating that most players position themselves at both the center and the four corners of the courts when preparing to attack. On the contrary, when defending, players intend to stay predominantly in the center of the court to enable quick reactions to any type of shot.

Table \ref{tab:shot_type} presents the numbers of different strike orders by shot types. 
It is evident that with the exception of the Clear and Smash shots, other shot types do not target around the head area.
Furthermore, more than half of the Push/Rush, Defensive Shot, and Short Service shots are executed with the backhand, while other shot types predominantly utilize the forehand.
Notably, there is a significant disparity in hitting position based on the service type, with almost all short services directed below the net and almost all long services aimed above the net.
% the service type (i.e., how the player starts the first shot) has the most distinct difference in hitting position, with almost all short services hitting below the net and almost all long services hitting above the net.

\vspace{-1pt}
\subsection{Evaluation Metrics}
The evaluation scenarios will be assessed by cross-entropy for shot type prediction and mean absolute error (MAE) for area coordinates prediction, similar to the original work.
Given the stochastic nature of this task, each team will be required to generate 6 predicted sequences for each rally, from which the closest one to the ground truth will be selected for evaluation.
It is worth noting that the original work on the stroke forecasting task involved generating 10 sequences, a number we have reduced to 6 for efficiency.

\vspace{-1pt}
\subsection{Official Baseline}
Generally, ShuttleNet \cite{DBLP:conf/aaai/WangSCP22} is a position-aware fusion of rally progress and player styles framework consisting of Transformer-based architectures.
ShuttleNet has demonstrated superior performance in predicting the next strokes compared to conventional sequential.
This is attributed to its turn-based architecture, which separates the styles of both players in a rally and integrates them with the current rally condition.

\begin{table}
    \centering
    \small
    \caption{Performance of the stroke forecasting task in CoachAI Badminton Challenge 2023 (Track 2).}
    \label{tab:performance}
    \scalebox{0.90}{\begin{tabular}{cc|cc|c}
    \toprule
    % & \multicolumn{4}{c}{$\tau=4$}\\
    % \cmidrule{1-5}
    \textbf{Rank} & \textbf{Team} & \textbf{CE} & \textbf{MAE} & \textbf{Total} \\
    \midrule
    1 & Intro\_to\_AI\_team8 & \textbf{1.7892} & 0.7884 & \textbf{2.5776} \\
   \midrule
    2 & Badminseok & 1.8127 & 0.7703 & 2.5830 \\
    \midrule
    3 & NYCU-group4 & 1.8411 & 0.7826 & 2.6237 \\
    \midrule
    4 & YHY & 1.9685 & \textbf{0.6797} & 2.6482 \\
    \midrule
    5 & 20 & 1.9366 & 0.7490 & 2.6856 \\
    \midrule
    6 & LOL & 1.9390 & 0.7618 & 2.7008 \\
    \midrule
    7 & AI Project \#15 & 1.9536 & 0.7483 & 2.7019 \\
    \midrule
    8 & Group27 & 2.0200 & 0.7139 & 2.7339 \\
    \midrule
    9 & Team\_13 & 1.9681 & 0.7671 & 2.7352 \\
    \midrule
    10 & LinDan & 2.0097 & 0.7743 & 2.7841 \\
    \midrule
    11 & Intro\_to\_AI\_group\_5 & 1.9710 & 0.8726 & 2.8436 \\
    \midrule
    12 & 14 & 2.1718 & 0.7013 & 2.8731 \\
    \midrule
    \midrule
    - & ShuttleNet \cite{DBLP:conf/aaai/WangSCP22} & 2.1777 & 0.6997 & 2.8774 \\
    \midrule
    \midrule
    13 & GD\_Wang & 2.2479 & 0.7081 & 2.9560 \\
    \midrule
    14 & ShuttleFold & 2.1277 & 1.0107 & 3.1385 \\
    \midrule
    15 & Awesome Badminton & 2.6579 & 0.7120 & 3.3699 \\
    \midrule
    16 & Badminton is all you need & 2.6579 & 0.7120 & 3.3699 \\
    \midrule
    17 & Group\_28 & 4.6615 & 1.0268 & 5.6883 \\
    \bottomrule
\end{tabular}}
\end{table}

\vspace{-1pt}
\section{Participating System}
Approximately 95 participants have joined the challenge system, with 16 teams submitting their testing results for the final phase of the CoachAI Badminton Challenge 2023 (Track 2).
Most teams modified the official baseline, ShuttleNet, to address the task.
The brief descriptions of the submitted methods are summarized in the video due to the page limit.

\section{Results and Findings}

Table \ref{tab:performance} reports the official leaderboard of the participants' methods.
It is observed that Team Intro\_to\_AI\_team 8 slightly outperforms the competition Team Badminseok, while Team YHY demonstrates the best performance in terms of area predictions.
11 teams perform better than ShuttleNet; however, we notice that these methods are built on top of ShuttleNet, which highlights the flexibility of ShuttleNet and the potential for improvements in certain aspects, e.g., hyper-parameters and activation functions.
In addition, the improvements made by the participating teams primarily focus on the shot type prediction (2.1777 $\rightarrow$ 1.7892); however, the performance of area coordinates only marginally surpasses ShuttleNet (0.6997 $\rightarrow$ 0.6797), with most teams being inferior to the baseline.
This underscores the challenge of effectively integrating two predictions, which may be a potential future direction for exploration.
\vspace{-3pt}
\section{Conclusion}
In this paper, we propose ShuttleSet22, an extended dataset from ShuttleSet with stroke-level badminton singles records.
ShuttleSet22 consists of fine-grained metadata to reinforce researchers to explore various aspects and derive insightful findings.
Exploratory data analysis is conducted to elucidate the fundamental compositions of the dataset, aiming to bridge the gap between researchers outside the badminton domain.
To foster researchers to incorporate advanced techniques into badminton analytics, we introduce Track 2, a challenge within CoachAI Badminton Challenge 2023, focused on enhancing the effectiveness of stroke forecasting.
In addition, we establish the state-of-the-art baseline for the task.
We summarize and discuss the methods and insights from participants to provide potential avenues for future research improvement.

\bibliographystyle{named}
\bibliography{ijcai24}

\begin{thebibliography}{}

\bibitem[\protect\citeauthoryear{Ban \bgroup \em et al.\egroup }{2022}]{DBLP:conf/mm/BanSAL22}
Kar{-}Weng Ban, John See, Junaidi Abdullah, and Yuen~Peng Loh.
\newblock Badmintondb: {A} badminton dataset for player-specific match analysis and prediction.
\newblock In {\em MMSports@MM}, pages 47--54. {ACM}, 2022.

\bibitem[\protect\citeauthoryear{Cartron}{2022}]{shuttlecock-cqzy3_dataset}
Mathieu Cartron.
\newblock Shuttlecock dataset.
\newblock \url{ https://universe.roboflow.com/mathieu-cartron/shuttlecock-cqzy3 }, mar 2022.
\newblock visited on 2023-05-06.

\bibitem[\protect\citeauthoryear{Chang \bgroup \em et al.\egroup }{2023}]{DBLP:journals/corr/abs-2211-12217}
Kai{-}Shiang Chang, Wei{-}Yao Wang, and Wen{-}Chih Peng.
\newblock Where will players move next? dynamic graphs and hierarchical fusion for movement forecasting in badminton.
\newblock In {\em {AAAI}}, pages 6998--7005. {AAAI} Press, 2023.

\bibitem[\protect\citeauthoryear{Chu and Situmeang}{2017}]{DBLP:conf/mir/ChuS17}
Wei{-}Ta Chu and Samuel Situmeang.
\newblock Badminton video analysis based on spatiotemporal and stroke features.
\newblock In {\em {ICMR}}, pages 448--451. {ACM}, 2017.

\bibitem[\protect\citeauthoryear{Decroos \bgroup \em et al.\egroup }{2019}]{DBLP:conf/kdd/DecroosBHD19}
Tom Decroos, Lotte Bransen, Jan~Van Haaren, and Jesse Davis.
\newblock Actions speak louder than goals: Valuing player actions in soccer.
\newblock In {\em {KDD}}, pages 1851--1861. {ACM}, 2019.

\bibitem[\protect\citeauthoryear{Huang \bgroup \em et al.\egroup }{2022}]{DBLP:conf/apnoms/HuangHLIW22}
Yu{-}Hsien Huang, Yung{-}Chang Huang, Hao~Syuan Lee, Ts{\`{\i}}{-}U{\'{\i}} Ik, and Chih{-}Chuan Wang.
\newblock S\({}^{\mbox{2}}\)-labeling: Shot-by-shot microscopic badminton singles tactical dataset.
\newblock In {\em {APNOMS}}, pages 1--6. {IEEE}, 2022.

\bibitem[\protect\citeauthoryear{Kim \bgroup \em et al.\egroup }{2022}]{DBLP:conf/kdd/KimKCYK22}
Hyunsung Kim, Bit Kim, Dongwook Chung, Jinsung Yoon, and Sang{-}Ki Ko.
\newblock Soccercpd: Formation and role change-point detection in soccer matches using spatiotemporal tracking data.
\newblock In {\em {KDD}}, pages 3146--3156. {ACM}, 2022.

\bibitem[\protect\citeauthoryear{Merhej \bgroup \em et al.\egroup }{2021}]{DBLP:conf/kdd/MerhejBMR21}
Charbel Merhej, Ryan~J. Beal, Tim Matthews, and Sarvapali~D. Ramchurn.
\newblock What happened next? using deep learning to value defensive actions in football event-data.
\newblock In {\em {KDD}}, pages 3394--3403. {ACM}, 2021.

\bibitem[\protect\citeauthoryear{Wang \bgroup \em et al.\egroup }{2020}]{Wang2020badminton}
Wei-Yao Wang, Kai-Shiang Chang, Teng-Fong Chen, Chih-Chuan Wang, Wen-Chih Peng, and Chih-Wei Yi.
\newblock Badminton coach ai: A badminton match data analysis platform based on deep learning.
\newblock {\em Physical Education Journal}, 53(2):201--213, 2020.

\bibitem[\protect\citeauthoryear{Wang \bgroup \em et al.\egroup }{2021}]{DBLP:conf/icdm/WangCYWFP21}
Wei{-}Yao Wang, Teng{-}Fong Chan, Hui{-}Kuo Yang, Chih{-}Chuan Wang, Yao{-}Chung Fan, and Wen{-}Chih Peng.
\newblock Exploring the long short-term dependencies to infer shot influence in badminton matches.
\newblock In {\em {ICDM}}, pages 1397--1402. {IEEE}, 2021.

\bibitem[\protect\citeauthoryear{Wang \bgroup \em et al.\egroup }{2022a}]{10.1145/3551391}
Wei-Yao Wang, Teng-Fong Chan, Wen-Chih Peng, Hui-Kuo Yang, Chih-Chuan Wang, and Yao-Chung Fan.
\newblock How is the stroke? inferring shot influence in badminton matches via long short-term dependencies.
\newblock 14(1), nov 2022.

\bibitem[\protect\citeauthoryear{Wang \bgroup \em et al.\egroup }{2022b}]{DBLP:conf/aaai/WangSCP22}
Wei{-}Yao Wang, Hong{-}Han Shuai, Kai{-}Shiang Chang, and Wen{-}Chih Peng.
\newblock Shuttlenet: Position-aware fusion of rally progress and player styles for stroke forecasting in badminton.
\newblock In {\em {AAAI}}, pages 4219--4227. {AAAI} Press, 2022.

\bibitem[\protect\citeauthoryear{Wang \bgroup \em et al.\egroup }{2023}]{ShuttleSet}
Wei{-}Yao Wang, Yung{-}Chang Huang, Tsi{-}Ui Ik, and Wen{-}Chih Peng.
\newblock Shuttleset: A human-annotated stroke-level singles dataset for badminton tactical analysis.
\newblock In {\em {KDD}}. {ACM}, 2023.

\bibitem[\protect\citeauthoryear{Wang}{2022}]{DBLP:conf/cikm/Wang22}
Wei{-}Yao Wang.
\newblock Modeling turn-based sequences for player tactic applications in badminton matches.
\newblock In {\em {CIKM}}, pages 5128--5131. {ACM}, 2022.

\end{thebibliography}

\end{document}